# A Hybrid Algorithm Based Robust Big Data Clustering for Solving Unhealthy Initialization, Dynamic Centroid Selection and Empty clustering Problems with Analysis

Y. A. Joarder[1] and Mossabir Ahamed[2]

[1,2] Department of Computer Science and Engineering, World University of Bangladesh (WUB),
Dhaka, Bangadesh

**Abstract:** Big Data is a massive volume of both structured and unstructured data that is too large and it also difficult to process using traditional techniques. Clustering algorithms have developed as a powerful learning tool that can exactly analyze the volume of data that produced by modern applications. Clustering in data mining is the grouping of a particular set of objects based on their characteristics. The main aim of clustering is to classified data into clusters such that objects are grouped in the same clusters when they are corresponding according to similarities and features mainly. Till now, K-MEANS is the best utilized calculation connected in a wide scope of zones to recognize gatherings where cluster separations are a lot than between gathering separations. Our developed algorithm works with K-MEANS for high quality clustering during clustering from big data. Our proposed algorithm "EG K-MEANS" : Extended Generation K-MEANS solves mainly three issues of K-MEANS: unhealthy initialization, dynamic centroid selection and empty clustering. It ensures the best way of preventing unhealthy initialization, dynamic centroid selection and empty clustering problems for getting high quality clustering.

## 1 Introduction

Big Data is a huge volume of both structured and unstructured data and clustering algorithms have developed as a powerful learning tool that can exactly analyze the volume of data. The main aim of clustering is to classified data into high quality clusters. K-MEANS is the best one that can do high quality clustering. As a delegate based bunching approach, K-MEANS offers an amazingly proficient inclination plummet approach to the complete squared blunder of portrayal, but, it is not just requests the parameter K, yet it additionally makes suspicions about the comparability of thickness among the groups. Subsequently, it is significantly influenced by commotion. Suppose to be more genuinely, it can frequently be pulled in to nearby optima, but, its drenching in a different conspires. We have presented a powerful hereditary calculation that consolidates the limit of hereditary administrators to combine various arrangements of the inquiry space with the abuse of the slope climber. The finding capability of genetic algorithms is exploited in order to search for appropriate cluster centres in the feature space such that a similarity metric of the resulting clusters is optimized. The chromosomes, which are represented as strings of real numbers, encode the centers of a fixed number of clusters. The superiority of the Genetic Algorithm clustering algorithm over the commonly used K-MEANS algorithm is extensively demonstrated for different real-life data sets. One of the most difficult tasks as per need of the user is to extract relevant information. Different terms of data mining are used for offering the needed information in such way that it can be parallel easily by the users. Several techniques of data mining are used like clustering, classifications which help to find out the hidden information in an independent way. Clustering is used to group similar objects into clusters so that same type of information can be restored easily by the humans. It is widely considered to be an important step in data exploration where interesting patterns and structures that reside in the dataset are extracted in spite of having negligible background knowledge. Different clustering methods are like Partitioning method, Hierarchical method, Grid-based method, Density-based method, and Model-based methods are used that employ different techniques to group the objects accordingly so that relevant information can be captured easily. The proposed research uses different technique of K-MEANS clustering algorithm for arranging objects of similar kinds. The prescribed technique is used for reducing the steps for K-MEANS clustering algorithm by employing genetic algorithm. Empty clusters can happen when using K-MEANS clustering algorithm, if the random initialization is poor, the number of K is in appropriate, the number of K is more than the number of data points in the data set. The original K-MEANS algorithms is not designed to handle this situation. If we find empty clusters while running K-MEANS, it will drop those clusters in the next iteration. Therefore, we may end up with fewer final clusters than you initially gave to the algorithm. To avoid this problem, we want to try different K or improve initialization of the initial cluster centers.

The K-MEANS algorithm is one of the most widely used clustering algorithms and has been applied into different fields of science and technology. One of the vital problems of the K-MEANS algorithm is that it may produce empty clusters and outlier these are depending on initial center vectors. For static execution of the K-MEANS, this problem is considered insignificant and can be solved by executing the algorithm for several numbers of times. In these problem, where the K-MEANS algorithm is used for as an integral part of some the higher level application, this empty cluster problem and outlier may produce anomalous behavior of the system and may have lead to significant performance of degradation. This research presents a hybrid algorithm that works with the K-MEANS algorithm to efficiently eliminate unhealthy initialization, dynamic centroid selection and empty clustering problem.

We have proved that our proposed algorithm is semantically equivalent to the original K-MEANS and there is no performance degradation due to incorporated modification or changes. Results of the simulation in experiments using several data sets prove our claim. We also mentioned regarding on outlier. An outlier in a pattern is dissimilar with rest of the pattern into big data clustering. Outlier detection is an important issue in big data clustering. It has been used to detect and remove anomalous objects from data or data set. Outliers occur due to reasons of mechanical faults, changes in system behavior, fraudulent behavior, and human errors. Our algorithm ensures the methodology of detecting and removing outlier during clustering. In K-MEANS algorithm, clustering outliers are found by distance based approach and cluster based approach. The additional goal of the research is the outliers free big data clustering while no unhealthy initialization, dynamic centroid selection and empty clustering issues for making the clusters more reliable.

Our contributions:

- To develop a hybrid algorithm to eleminate unhealthy initialization, dynamic centroid selection and empty clustering during clustering from big data for high quality clustering
- To analize exprimental results of our proposed algorithm: EG KMEANS as well as compare the results with other realated algorithms.

## 2 Literature Review

Accomplishing robust clustering is a standout amongst the most notable issues in data mining. K-MEANS is by a long shot the most generally utilized clustering calculation. It combines decently fast, yet accomplishing a decent arrangement is not ensured. The clustering quality is exceptionally subject to the choice of the underlying centroid choices. In addition, when the quantity of groups expands, it begins to experience the effects of "empty clustering"[1]. Clustering evaluation is one of the most in many instances used information processing algorithms. K-MEANS stays the best popular clustering algorithm due to the fact of its simplicity [2]. There are few field of studies implemented on optimizing exceptional goals of K-MEANS algorithm where works Euclidean k-medians and geometric k-center [3]. Minimization of the whole of separations to the closest focus is the objective for Euclidean k-medians, and minimization of the most extreme separation from each point to its closest focus is the one for geometric k-focus variant. Another examination was done to look for a superior target capacity of K-MEANS [4]. Use of traditional partition based algorithms are limited to numeric data that works well for data with mixed numeric and categorical features [5]. K-MEANS algorithm is sensitive to the initial seeds (cluster centers) that will produces randomly. Bad initial seeds can easily lead K-MEANS to poor clustering in results [6]. Since, K-MEANS and KMEANS++ use the same hill-climbing approach, they obtain very rapidly to the local optima near the initial set $C_0$ of centers. Such local-optima have been reported in the literature as usually being of poor quality [7]. Due to the random selection of genetic algorithm, the chromosomes into the initial population may not be contain generation representing all clusters of the individual dataset. The crossover operation did not make new generation instead of it and only re-arranges the generation of parent chromosomes. The mutation operation slightly changes some genes and thereby in a way that creates new generation. However, the mutation operation typically performs operation such a small changes that new generations are still similar to original generation [8]. Due to the complex initial population selection into process, it suffers from a high complexity of $O(n^2)$ time. Surprisingly, many existing clustering techniques areusing genetic algorithms with an even worse complexity of $O(n^3)$ time [9]. Clustering are the records of a dataset element in such a way that similar individual records are clustering together in a cluster and dissimilar records are placed in different clusters. It has a broad range of applications in almost all areas including generation analysis [10]. Many genetic algorithms are selecting their initial population which may have an adverse impact on final clustering results. Clustering groups the records of a dataset in such a way that similar records are grouped together in a cluster and dissimilar records are placed in different clusters [11]. It has a broad range of applications in almost all areas including gene analysis [12]. A recent technique called GENCLUST uses an advanced approach in selecting its initial of population which was shown to be useful in achieving better clustering results [13][14]. The crossover operation is applied on the chromosomes of data set by finding pairs of parents. Data sets components are performed operation according to chromosome analysis of individual task [15]. The vocabulary tree defines a hierarchical quantization that is built by hierarchical K-MEANS clustering. A large set of representative descriptor vectors are used in the unsupervised training of the tree. Instead of K defining the final number of clusters or quantization cells, k defines the branch factor (number of children of each node) of the tree. An initial K-MEANS process is run on the training data, defining K cluster centers. The training data is then partitioned into K groups, where each group consists of the descriptor vectors' closest to a particular cluster center [16]. In the k-medians problem, we are given a set S of n points in a metric space and a positive integer S. The objective is to locate k medians among the points so that the sum of the distances from each point in S to its closest median is minimized [3].The intention is to share common understanding of the meaning of any term that has been used, and therefore it could support the database query tool to find functionally equivalent terms in cross-database search. In essence, this will improve retrieval consistency across resources and the recall and precision of the query result within resources [17]. High throughput data need to be processed, analyzed, and interpreted to address problems in life sciences. Bioinformatics, computational biology, and systems deal with biological problems using computational methods. Clustering is one of the methods used to gain insight into biological processes, particularly at the genomics level. Genetic clustering algorithms designed especially for analyzing gene expression data set [18]. K-MEANS algorithm is the most well-known and commonly used clustering method. It takes the input parameter, K, and partitions a set of n objects into K clusters so that the resulting intra-cluster similarity is high whereas the inter cluster similarity is low. Cluster similarity is measured according to the mean value of the objects in the cluster, which can be regarded as the clusters "center of gravity" [19]. Somewhere, for global optimization of complex functions. In the proposed FA, two types of explosion (search) processes are employed, and the mechanisms for keeping diversity of sparks are also well designed [20]. The genetic k-means clustering process to calculate a weight for each dimension in each cluster and use the weight values to identify the subsets of important dimensions that categorize different clusters. This is achieved by including the weight entropy in the objective function that is minimized in the k-means clustering process [21]. GENCLUST carefully selects high quality initial population which was experimentally shown to be effective. While clustering techniques based on genetic algorithms. GENCLUST has the following drawbacks. First, its initial population selection technique has a time complexity of $O(n^2)$ which can be problematic for big [22]. On the other hand, GENCLUST++ combines the best features of the fast hill-climbing of K-MEANS and KMEANS++, with the exploratory nature of genetic algorithms. The idea is that any solution of poor quality can rapidly be polished to a local optima (or nearby) using a few iterations of the K-MEANS' hill-climbing; however, combinations of one or more local-optima can only be achieved by genetic operators. To this end, GENCLUST++ incorporates a regular intervention (every 10 generations) where chromosomes that represent clustering solutions are polished by K-MEANS' hill-climber ensuring that the population has a large number of high performing chromosomes in the population. This benefits the genetic search, since operators like mutation and crossover only improve the fitness stochastically and with low probability [23].

## 3 Proposed algorithm "*Extended Generation K-MEANS (EG K-MEANS)" Algorithms*

**Algorithm 1: Extended Generation K-MEANS (EG K-MEANS)**

**Input** : Input data, *W*

**Output** : A set of clusters *X*

Input data, *W*

$D_0$ ← reduced data using SVM (*W*)

$T_s$ = Total number of data points ($D_0$)

/* First phase */

```
if float(T_s) == True:
        if (|T_d| < 10):
                H = T_d * A_c
        else:
                T_d
        D_1 ← round_floor(Td | H)
else:
        S ← insertion_sort(T_s | D_1)
/* Second Phase */
if S_i == S_i++:
        d ← eliminate_data_set(S)
else:
        S
/* Third phase */
if check_even(d | S) == True:
        d_even ← sort_even(d, 'asc')
        d_small = get_smallest(d_even)
        d_even = R_s / d_small
else:
        d_odd ← sort_odd(d, 'asc')
        d_small = get_smallest(d_odd)
        d_odd = R_s / d_small
/* Fourth phase */
```

**if** ($d_{even}$ == $d_{odd}$):

$K_f$ ← find_common_smallest($d_{even}$ , $d_{odd}$)

**else**:

$K_f$ ← get_small_value($d_{odd}$)

/* Cluster size (floor) */

C = $T_s$ / $K_f$

C = floor(C)

C = reverse(C)

/* P = optional parameter selects the position */

N = get_smallest($d_{even}$ , p)

/* Getting mean value from cluster(first parameter) using N = $d_{even}$(p) as index (second parameter) */

Y = mean(C, N)

/* Euclidean distance function */

X = euclidean(Y)

**return** *X*

---

### 3.1 Mathematical calculation of "*Extended Generation K-MEANS (EG K-MEANS)" Algorithms*

*Input data, (raw data), W*
*Data dimension reduction using nonlinear SVM*

$$T_s \text{ is the total number of data points (for example : we have overall 10 number data points)}$$

$T_d$ *is the original data set*
*(Therefore, there are 10 inividule data points, such as :*

$T_{d1}, T_{d2}, T_{d3}, T_{d4}, T_{d5}, T_{d6}, T_{d7}, T_{d8}, T_{d9}, T_{d10} \dots\dots\dots. T_{dn}$; *Where,* $T_{d1} = 1, T_{d2} = 2, T_{d3} = 3, T_{d4} = 4, T_{d5} = 5, T_{d6} = 6, T_{d7} = 7, T_{d8} = 8, T_{d9} = 9, T_{d10} = 10,\ \dots\dots\dots\dots\dots.. T_{dn} = z$*)*

*if (|[Td]|< 10)*

$$H = (Td * Ac)$$

*Where, Ac = Arbitrary Constant, which must be multiple of 10*

$$d_i = \left\lceil \frac{T_s}{i} \right\rceil \ where\ i = 2$$

$$d_i = \begin{matrix} d_i \neq d_{i-2} \\ R \\ i = 4 \\ i \bmod 2 = 0 \end{matrix} \qquad \left\lceil \frac{T_s}{i} \right\rceil$$

$$d_{even} = \{d_2, d_4, d_6, \dots\dots\dots, d_m\}\ where\ m = 2 \times n, when\ n\ is\ an\ integer$$

$$d_j = \left\lceil \frac{T_s}{j} \right\rceil \ where\ j = 3$$

$$d_j = \begin{matrix} d_j \neq d_{j-2} \\ R \\ j = 4 \\ j \bmod 2 = 1 \end{matrix} \qquad \left\lceil \frac{T_s}{j} \right\rceil$$

$$d_{odd} = \{d_3, d_5, d_7, \dots\dots\dots, d_m\}\ where\ m = 2 \times n + 1, when\ n\ is\ an\ integer$$

$$K = d_{even} \cap d_{odd}$$

$$K_f = min(K) \; if \; K \neq \{\}$$

*Or*

$$K_f = min(d_{odd}) \; if \; K = \{\}$$

$$D_{i,j} = \underset{i=1}{\overset{i=K_f}{R}} \; \underset{j=1+\left(\left\lceil T_s/K_f \right\rceil\right)\times(i-1)}{\overset{j=\left(\left(\left\lceil T_s/K_f \right\rceil\right)\times i\right)\leq T_s}{R}} \; T_d(j)$$

$$d_{even} = sort(d_{even}) \; ascending \; order$$

$$RD_{i,j} = reverse(D_{i,j})$$

$$M_i = \underset{j=d_{even}(p)}{\underset{i=1}{\overset{i=K_f}{R}}} \; RD_{i,j}$$

*Where p is the second position and* $\boldsymbol{d_{even}}(\boldsymbol{p})$ *is the second smallest element in the sorted array* $\boldsymbol{d_{even}}$.

$$F_{C_k} = R_{i=1}^{i=kj} \, R_{k=1}^{k=kj} \left\{RD_{i,j} : arg\,min\,dis(M_k, RD_{i,j}); \; \forall j\right\}$$

$$F_{C_k} \rightarrow Final\ Cluster$$

*Where,* $\boldsymbol{k}$ *is the number of cluster.*

### 3.2 The flowchart of "*Extended Generation K-MEANS (EG K-MEANS)" Algorithms*

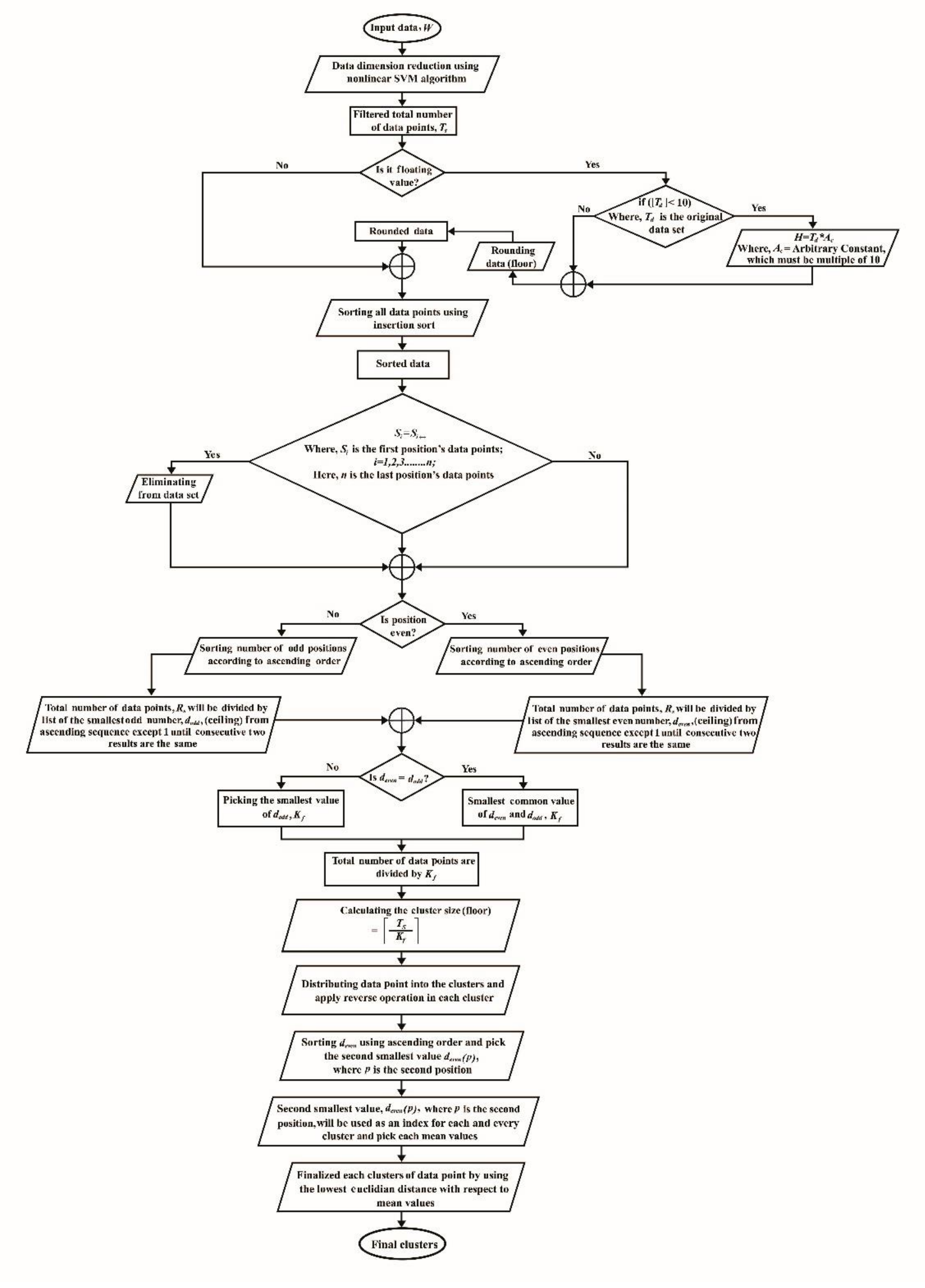

### 3.3 Explanation of the flowchart "*Extended Generation K-MEANS (EG K-MEANS)" Algorithms*

Here, we have given input some raw data (*W*) for our calculation. We have used Non Linear Support Vector Machines (SVM) algorithm for personalized raw data as our corresponding equation's input method system. Then filtered total number of data points, here, we have denoted this $T_{s.}$ Detecting floating value if found then accommodate equation where, $T_d$ has the original data set. Finalized step through rounding value using floor value. Rounded data pints has been sorting using insertion sort also assemble as ascending sequence. Separate the sequence of even and odd position of data points also sorting particular as ascending order. Denoted here, $R_s$ as total number of data points, $R_s$ has been divided by the smallest number of odd and even number denoted result sequenced as $d_{odd}$ and $d_{even}$ . Both result have been counting using celling value and from ascending sequence except 1 until consecutive two results are same. Number cluster denoting $K_f$ , If any similar value found between $d_{odd}$ and $d_{even}$ register then pick the smallest number of value if not then pick the smallest value of $d_{odd}$ . For denoting cluster size $R_s$ has divided by $K_f$ and result has been counted using floor value. Distributing data points has been into the clusters and apply reverse operation in each cluster segment. Calculate the sorting $d_{even}$ has been using ascending order and pick the second smallest value (position) $d_{even}(P)$. Finalized the each clusters of data points have been using smallest Euclidian distance with respect as mean values.

## 5 Results

In this section, we have reported the results of several experiments carried out to validate the merit of the proposed Extended Generation KMEANS (EG KMEANS). In addition, we have described the experimental setup that used to validate the proposed techniques with the results of the experiments to evaluate the performance of all below mentioned algorithms respectively.

**1. Examining data points for clustering**

[A] = {2, 4.3, 5, 6, 8, 9, 10, 90, 12, 21, 34}

[B] = {9, 80, 31, 15, 4, 8, 7, 90, 11}

[C] = {20, 3, 45, 26, 3, 2, 10, 8, 10, 3, 13}

[D] = {2.0, 4, 3, 5, 6, 8, 9, 10, 90, 12, 21, 34}

[E] = {3, 10, 15, 26, 18, 4, 1,-1}

[F] = {32, 34, 3, 15, 4, 8, 19, 32, 21}

[G] = {20, 9, 30, 15, 16, 98, 9, 10, 90}

Table-1: Compare between different cluster algorithms with some data points

| Data points | EG K-MEANS | GenClust++ | GenClust-F | AGCUK | GenClust-H | GAGR | K-MEANS |
|---|---|---|---|---|---|---|---|
| [A] | 0.911 | 0.92 | 1.512 | 1.690 | 1.311 | 1.32151 | 1.6212 |
| [B] | 0.99 | 0.991 | 1.009 | 1.452 | 1.110 | 1.321 | 1.60 |
| [C] | 0.85 | 0.821 | 1.002 | 1.034 | 1.109 | 1.453 | 1.034 |
| [D] | 0.812 | 0.730 | 1.213 | 1.321 | 1.490 | 1.321 | 1.321 |
| [E] | 0.714 | 0.713 | 1.12 | 1.60 | 1.54 | 1.31 | 1.321 |
| [F] | 0.721 | 0.723 | 1.67 | 1.94 | 1.42 | 1.21 | 1.321 |
| [G] | 0.91 | 0.992 | 1.42 | 1.942 | 1.23 | 1.92 | 1.43 |

We have analized several clustering algorithm by different data points on MATLAB. In addition, we have found random result value during clustering. The above result table has showed the best result of time efficient during clustering with mentioned data points set. Here, we have observed that most efficient result for data point sets [A], [B], [F] and [G]. Obtained results: 0.911, 0.99, 0.721 and 0.91, those are very near distance then GenClust++ and more efficient then K-MEANS.

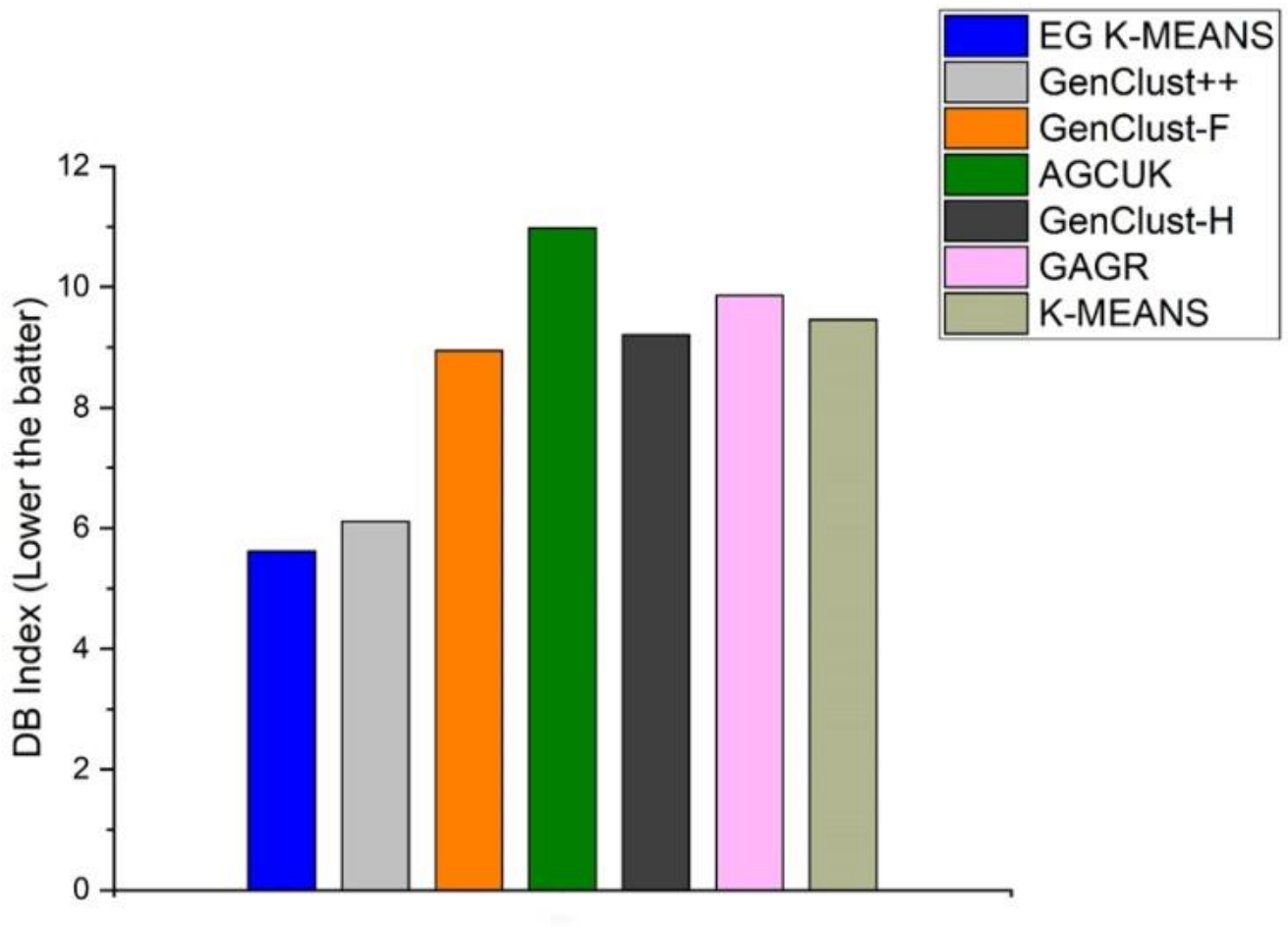

Fig 10: Average DB Index (Lower Number)

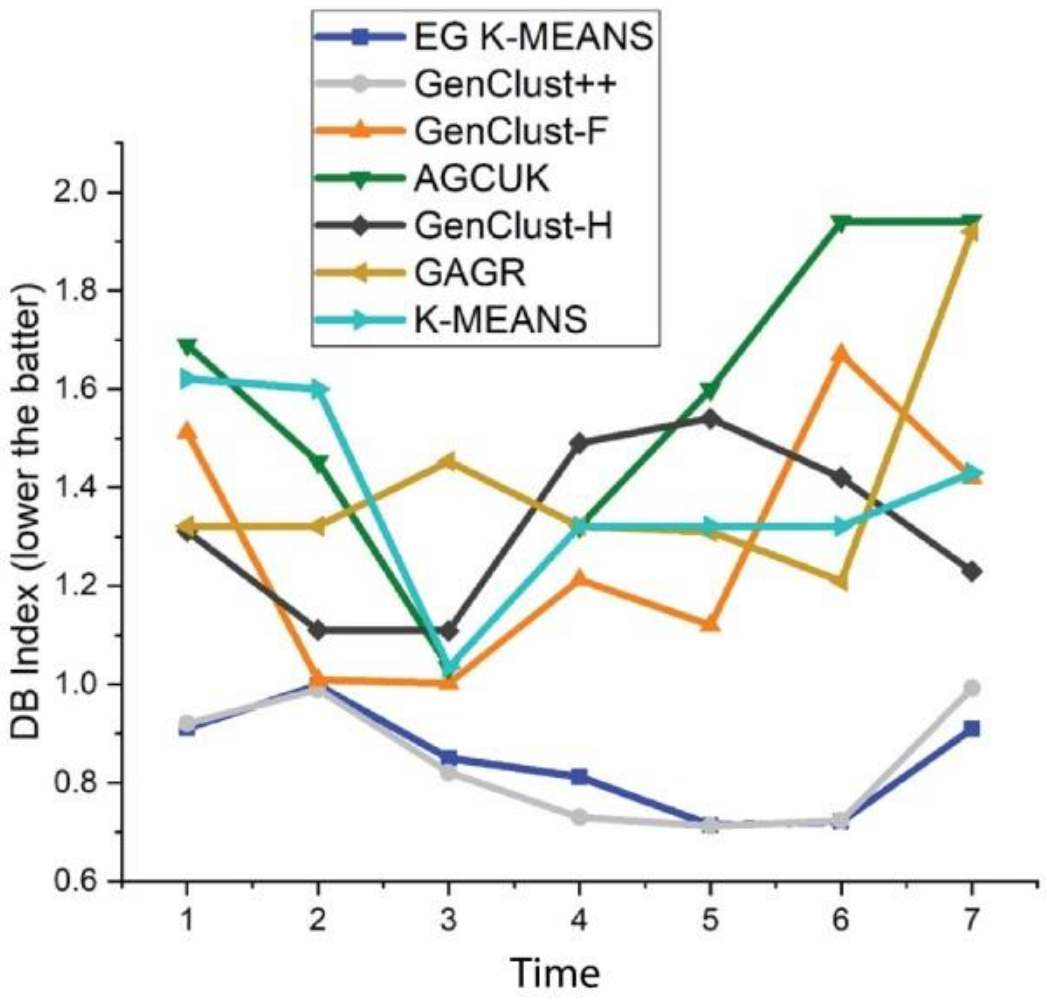

Fig 11: Average DB Index (Lower Number) with each data points

Here, we have seen that our EG K-MEANS get value less minimum value during clustering on big data. Our proposed algorithm have been worked for minimizing clustering problem also eliminating unhealthy initialization, dynamic centroid selection and empty clustering problems during clustering from big data. Bold value into the result table is efficient for clustering any data points.

**2 Examining datasets:**

We have taken 10 publicly available datasets in our experiment from the University of California at Irvine (UCI) repository, the Jin Genomics Datasets (JGD) repository and Kaggle datasets. Each feature of each dataset has been normalized so as to have zero mean and unit standard deviation. The details of these 10 datasets are listed in Table 2.

Table 2: A brief introduction to the datasets

| Datasets | Instances | Instances without Missing value | Attributes | Classes | Repository |
|---|---|---|---|---|---|
| Page Blocks Classification | 5472 | 5472 | 10 | 5 | KAGGLE |
| Mammographic Mass | 961 | 830 | 5 | 2 | KAGGLE |
| Credit Approval | 345 | 345 | 6 | 1 | KAGGLE |
| Yeast | 1484 | 1484 | 8 | 10 | JGD |
| Glass Identification | 214 | 214 | 10 | 7 | JGD |
| Liver Disorder | 345 | 345 | 6 | 2 | UCI |
| Dermatology | 366 | 358 | 34 | 6 | UCI |
| Haberman | 306 | 306 | 3 | 3 | UCI |
| Tic-Tac-Toe | 958 | 958 | 9 | 2 | UCI |
| Contraceptive Method Choice | 1473 | 1473 | 2 | 3 | UCI |

**3. DB Index (lower the better) of the techniques on the 10 datasets.**

Table 3: Compare between Different cluster algorithms with DB Index value

| Datasets | Extend Generation KMEANS | GenClust++ | GenClust-H | GenClust-F | AGCUK | GAGR | K-MEANS |
|---|---|---|---|---|---|---|---|
| Page Blocks Classification | 0.724 | 0.9300 | 0.9699 | 1.3005 | 1.0220 | 1.7854 | 1.1902 |
| Mammographic Mass | 0.326 | 0.5814 | 0.6067 | 0.6076 | 0.7918 | 1.4228 | 1.3025 |
| Credit Approval | 0.45 | 0.4578 | 1.2177 | 1.8548 | 0.977 | 1.255 | 1.645 |
| Yeast | 1.013 | 1.5461 | 1.5944 | 1.8663 | 1.6315 | 1.9571 | 1.7531 |
| Glass Identification | 1.395 | 1.1978 | 1.5079 | 1.4221 | 1.4563 | 1.3367 | 1.3963 |
| Liver Disorder | 0.987 | 1.1047 | 1.1295 | 1.4352 | 1.2080 | 1.7698 | 1.6065 |
| Dermatology | 0.765 | 1.1155 | 1.1759 | 1.4950 | 1.2307 | 2.1240 | 2.1687 |
| Haberman | 1.058 | 1.3380 | 1.3694 | 1.3449 | 1.7001 | 1.3811 | 3.0937 |
| Tic-Tac-Toe | 1.001 | 0.6639 | 0.7220 | 0.7318 | 0.987 | 1.123 | 1.321 |
| Contraceptive Method Choice | 1.311 | 1.0246 | 1.0246 | 1.0262 | 1.026 | 1.524 | 1.224 |

In the table and graph has been represented the accuracy rate of best features by Extended Generation K-MEANS using Imputation method with K-MEANS classification algorithm. In this phase, we compare between many Imputation methods. We have seen this graph is mostly probable. Some of the datasets have been given best result.

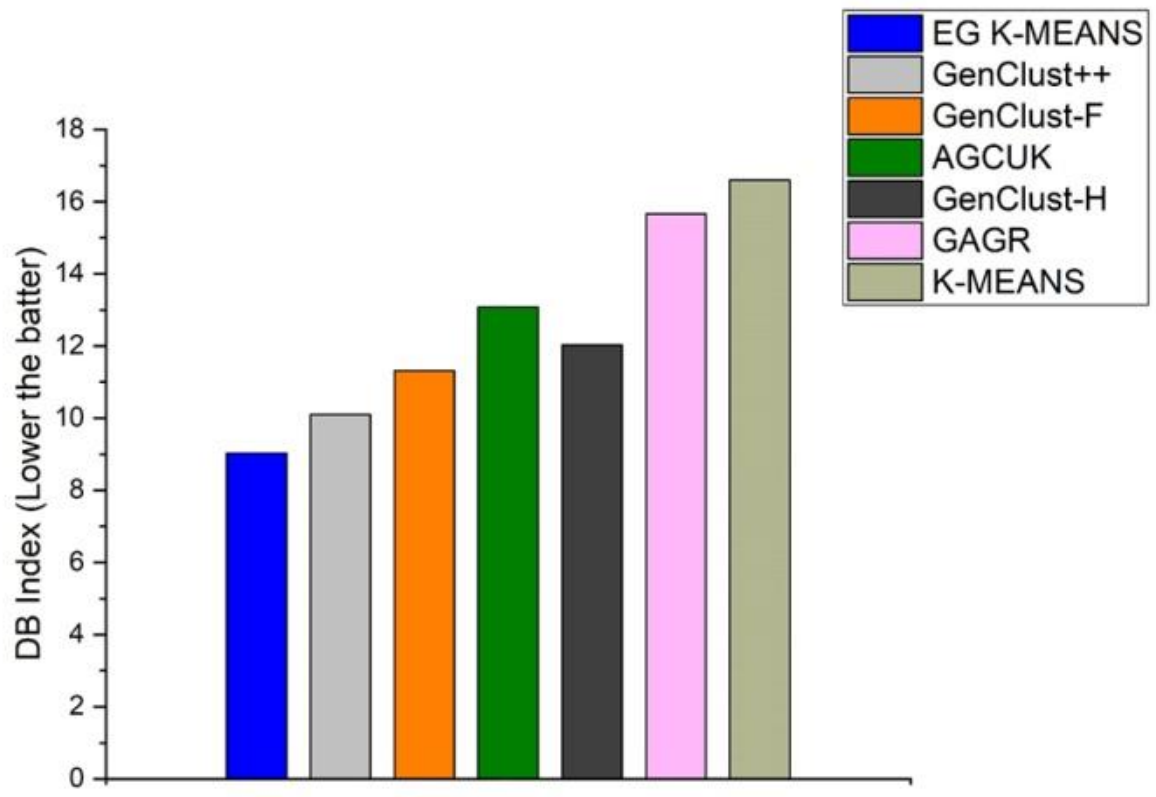

Fig 12: Average DB Index in datasets

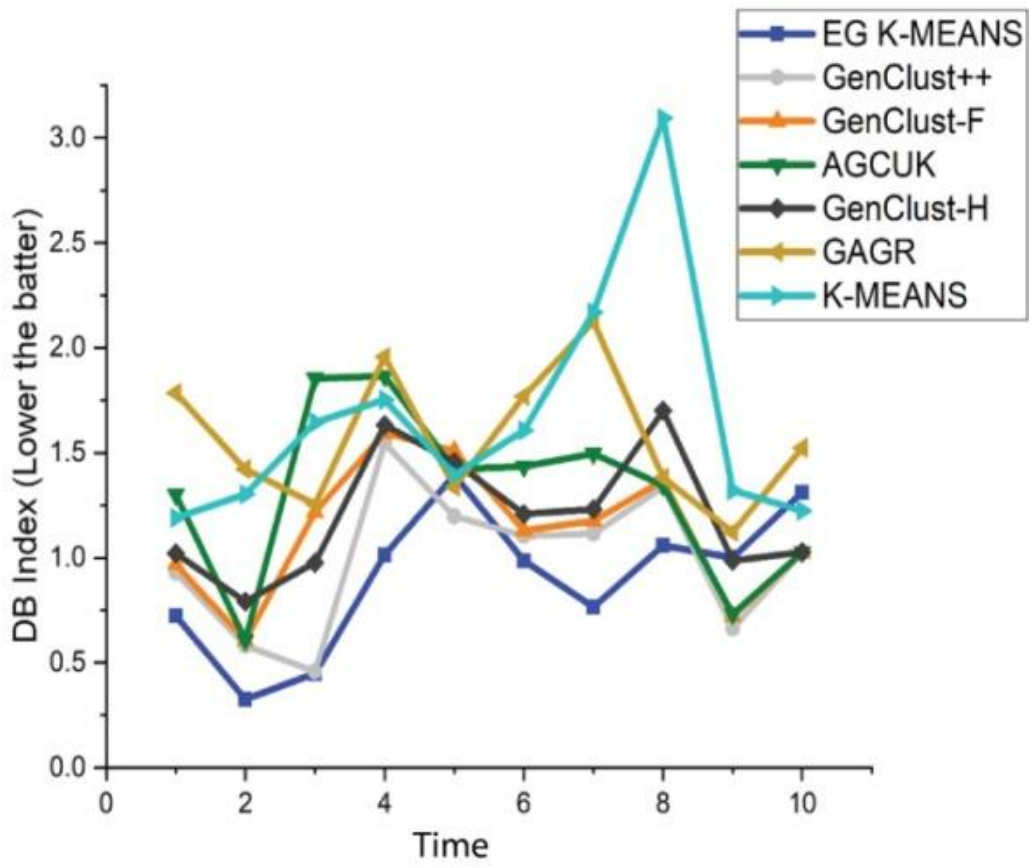

Fig 13: DB Index in datasets with each value

Since, the clustering solutions obtained by the techniques can vary between different runs, we have run each technique ten (10) times on each dataset recording their cluster quality. The tables have been presented the average results of the ten (10) clustering solutions on each dataset for each technique. In contrast to the tables, we have used the figures to present the average results over all datasets and over all repetitions for a dataset. The discussion of the figures will be followed by a discussion of the statistical significance analysis we have performed through the non-parametric sign test analysis.

## 6 Conclusion

In this research, we haved propose to use the EG K-MEANS measure as a viable alternative to imputation and marginalization approaches to handle the problem of unhealthy initialization, dynamic centroid selection and empty clustering in big data clustering. Clustering in data mining is the grouping of a particular set of objects based on their characteristics, aggregating them according to their similarities. Our proposed algorithm "EG K-MEANS": Extended Generation K-MEANS has been removed unhealthy initialization, dynamic centroid selection and empty clustering from big data for high quality clustering.Our proposed algorithm is semantically equivalent to the original K-MEANS and there is no performance degradation due to incorporatedmodification or changes. Results of the simulation in experiments using several data sets prove our claim. In future, we will try to minimise the time complexity and improve the accuracy rate.